\documentclass{article}

\usepackage{microtype}
\usepackage{graphicx}
\usepackage{subfigure}
\usepackage{booktabs} 

\usepackage{times}

\usepackage{soul}
\usepackage{url}
\usepackage[hidelinks]{hyperref}
\usepackage[utf8]{inputenc}
\usepackage[small]{caption}
\usepackage{graphicx}
\usepackage{amsmath}
\usepackage{booktabs}
\usepackage{multirow}
\usepackage{bm}
\usepackage{amsfonts}

\newtheorem{theorem}{Theorem}

\usepackage{hyperref}

\usepackage[accepted]{icml2021}

\icmltitlerunning{RSGP}

\begin{document}

\twocolumn[
\icmltitle{Regularization of Sparse Gaussian Processes with Application to Latent Variable Models}



\icmlsetsymbol{equal}{*}

\begin{icmlauthorlist}
\icmlauthor{Rui Meng}{lbl}
\icmlauthor{Herbert Lee}{ucsc}
\icmlauthor{Braden Soper}{llnl}
\icmlauthor{Priyadip Ray}{llnl}
\end{icmlauthorlist}

\icmlaffiliation{lbl}{Lawrence Berkeley National Laboratory}
\icmlaffiliation{ucsc}{University of California, Santa Cruz}
\icmlaffiliation{llnl}{Lawrence Livermore National Laboratory}

\icmlcorrespondingauthor{Rui Meng}{rmeng@lbl.gov}

\icmlkeywords{Machine Learning, ICML}

\vskip 0.3in
]



\printAffiliationsAndNotice{\icmlEqualContribution} 

\begin{abstract}
Gaussian processes are a flexible Bayesian nonparametric modelling approach that has been widely applied, but poses computational challenges. To address the poor scaling of exact inference methods, approximation methods based on sparse Gaussian processes (SGP) are attractive. An issue faced by SGP, especially in latent variable models, is the inefficient learning of the inducing inputs, which leads to poor model prediction. We propose a regularization approach by balancing the reconstruction performance of data and the  approximation performance of the model itself. This regularization improves both 
inference and prediction performance. We extend this regularization approach into latent variable models with SGPs and show that performing variational inference (VI) on those models is equivalent to performing VI on a related empirical Bayes model.
\end{abstract}

\section{Introduction} \label{sec:introduction}
A Gaussian process (GP) is a generalization of a multivariate Gaussian distribution that can be seen as a random process in the space of general continuous functions \cite{Rasmussen_2005}. Due to its flexibility, it is applied in a wide range of fields. 
However, exact inference is computationally expensive with time complexity $\mathcal{O}(N^3)$, where $N$ is the number of data points. This renders exact GP inference infeasible for large datasets. To address this issue, approximation algorithms are required. Many approximation algorithms employ a set of inducing points, leading to sparse Gaussian process (SGP) methods. To select the inducing points such approximate methods rely on heuristics  \cite{Lawrence_2003,Seeger_2003}, use pseudo targets obtained during the optimization of the log-marginal likelihood \cite{Snelson_2006}, or learn the inducing inputs during the optimization of the elbow lower bound of the log-marginal likelihood \cite{Titsias_2009,Hensman_2012}.

All of these SGP methods concentrate on the reconstruction performance in the training process by maximizing either the likelihood or a lower bound of the likelihood, but none of them exploit the difference of the likelihood functions between a full GP model and a certain approximate SGP model given the same model parameters, which is called the approximation performance. Thus we propose a regularization framework that balances the reconstruction performance and the approximation performance. We empirically show that our proposed regularization framework consistently contributes to better predictive performance. 

At the same time, learning inducing inputs during the optimization plays a big role for most sparse Gaussian process methods \cite{Snelson_2007,Titsias_2009,Hensman_2013} because it empirically contributes to better prediction results. However, due to the non-linearity and non-convexity of objective functions, even if we initialize inducing point locations with the K-means clustering method, those approaches tend to get stuck in local maxima and cause poor model prediction performance. This issue is recently mentioned in \cite{zhe2017regularized}. Two classes of approaches exist to relieve it. One is to put an informative structure on training inputs \cite{zhe2017regularized}, and the other is to model the inducing inputs \cite{hensman2015mcmc,rossi2020sparse}. However, our work leverages an 
information theoretic regularization term to mitigate the problem of local maxima where our regularization term is strictly convex to the summary statistics of inducing inputs. Specifically, by adjusting the weight of the regularization term, our framework balances the reconstruction performance and the degree of convexity of the objective function, allowing the regularized objective function 
to find
superior local maxima during the training process.

The Gaussian process latent variable model (GPLVM) is proposed by \cite{Lawrence_2003} as a probabilistic dimensionality reduction method. This method extends the linear mappings from the embedding space in dual probabilistic principal component analysis to nonlinear mappings, however it does not have good scaling properties. 
\cite{Titsias_2010} proposes Bayesian GPLVM, which incorporates inducing inputs into GPLVM to make the algorithm scalable. \cite{Hensman_2013} proposes stochastic variational inference on latent variable models based on Gaussian processes. 
We naturally extend our regularization framework into those latent variable models by replacing the data inputs by unknown embedding inputs. 

The main contributions of this work are: 1) We propose a regularization approach for inducing-point based SGP models. To the best of our knowledge, it is the first 
SGP method to explicitly 
consider the trade-off between the reconstruction performance and the approximation performance; (2) This regularization employs a convex divergence-based regularization term. Leveraging the benefits of this convexity, our framework 
is robust 
with respect to initialization, alleviating the problem of local maxima and improving the prediction performance; 3) The regularization is extended to latent variable models based around Gaussian processes, and our approach increases the generalization capability of latent variables to achieve better reconstruction performance. Also, we theoretically justify the use of this regularization term by proving that performing variational inference (VI) on a GPLVM with this regularization term is equivalent to directly performing VI on a related empirical Bayes model with a prior on its inducing inputs.

The remainder of this paper is organized as follows. In Section~\ref{sec:background}, we briefly introduce the literature of sparse Gaussian process (SGP) models. Next we provide the regularization framework of SGPs in Section~\ref{sec:RSGP} as well as its analysis. We empirically analyze various performances of our regularization framework on different SGP models in Section~\ref{sec:CB}. Moreover, we extend our framework into latent variable models and provide its analysis in Section~\ref{sec:RLSGP} and demonstrate the model performance on two real datasets in Section~\ref{sec:experiments}.

\section{Background of Sparse Gaussian Processes} \label{sec:background}
Suppose we have $N$ observations (i.e., responses) ${\bm y} = \{y_i\}_{i=1}^N$ with data inputs (i.e., explanatory variables) ${\bm X} = \{{\bm x}_i\}_{i=1}^N$, where each entry $y_i$ is a noisy observation of the function $f({\bm x}_i)$. We consider the noise to be independent Gaussian with with precision $\beta$. The latent function $f$ is modeled by a Gaussian process, and the vector $\bm f$ is composed of values of the function at the points $\bm X$. 
We introduce a set of inducing variables ${\bm u} = \{u_i\}_{i=1}^M$ and inducing inputs ${\bm Z} = \{{\bm z}_i\}_{i=1}^M$ such that $u_i = f({\bm z}_i)$ for $i=1,..,M$.
Then the SGP model is formulated as
\begin{align}
    p(\bm y| \bm f) & = \mathcal{N}(\bm y| \bm f, \beta^{-1}\bm I)\,, \nonumber\\
    p(\bm f| \bm u) & = \mathcal{N}(\bm f| \bm K_{nm} \bm K_{mm}^{-1} u, Q)\,, \nonumber \\
    p(\bm u) & = \mathcal{N}(\bm 0, \bm K_{mm})\,, \nonumber
\end{align}
where $\bm K_{mm}$ is the covariance function evaluated between all the inducing inputs,  $\bm K_{nm}$ is the covariance function evaluated between all the data inputs and inducing inputs, and $\bm Q = \bm K_{nm}\bm K_{mm}^{-1}\bm K_{mn}$. We also introduce the notation $\tilde{\bm K} = \bm K_{nn} - \bm Q$.


All inducing point GP methods avoid working directly with the full GP model by solving a related surrogate optimization problem. For example, sparse GP methods maximize the log marginal likelihood of a GP with a low-rank approximation to the full covariance matrix, while variational approaches maximize a lower bound to the evidence.   Regardless of approach we denote the objective function of the surrogate optimization problem by $\ell_1 = \ell_1(\bm y; \bm \theta)$, and we denote the log marginal likelihood of the full GP by $\ell_0 = \ell_0(\bm y; \bm \theta) = \log \mathcal{N}(\bm y|\bm 0, \bm K_{nn} + \beta^{-1} \bm I)$ for data $\bm y$ and GP parameters $\bm \theta$. 

In both the full and sparse GP model, inference involves finding the model parameters $\bm \theta_i$ that maximize the objective function, i.e., $\hat{\bm \theta}_i = \arg\max\limits_{\bm \theta} \ell_i(\bm y; \bm\theta)$, $i = 0, 1$. However, there is no theoretically guaranteed bound for the difference of the two objective functions, suggesting that the optimal model parameters from two models may be significantly different. This could lead to poor model fitting or generalization performance. In the next section we propose a new regularization approach that penalizes the surrogate objective function $\ell_1$ for diverging from the true objective function $\ell_0$. In this way the sparse GP model explicitly considers not just how well the surrogate model fits the data (reconstruction performance), but also how well the surrogate model approximates the full model (approximation performance). 

In the remainder we consider the following state-of-the-art inducing-point methods: subset of regression (SoR) \cite{quinonero2005unifying}, deterministic training conditional approximation (DTC) \cite{Seeger_2003}, fully independent training conditional approximation (FITC) \cite{Snelson_2006}, sparse Gaussian process regression (SGPR) \cite{Titsias_2009} and stochastic variational inference for sparse Gaussian process (SVGP) \cite{Hensman_2013}. Objective functions of these different models are summarized in Table~\ref{tab: SGP_objective function}, while other properties can be found in Appendix A in the supplementary materials.

\begin{table}[ht!]
	\caption{The objective functions in the training procedure for sparse Gaussian Process models.}
	\label{tab: SGP_objective function}
	\centering
	\resizebox{\linewidth}{!}{
	\begin{tabular}{c|c}
		\hline
		MODEL & $\ell_1$ \\
		\hline 
		SoR/DTC & $\log\left(\mathcal{N}(\bm y| \bm 0, \bm Q + \beta^{-1}\bm I)\right)$ \\
		\hline
		FITC & $\log\left(\mathcal{N}(\bm y| \bm 0, \bm Q + \mathrm{diag}(\bm \tilde{K})+\beta^{-1}\bm I)\right)$\\
		\hline
		SGPR & $\log \mathcal{N}(\bm y| \bm 0, Q + \beta^{-1}\bm I) -\frac{\beta}{2}\mathrm{tr}(\tilde{K})$\\
		\hline 
		SVGP & refer to (4) in \cite{Hensman_2013} \\
		\hline
	\end{tabular}
    }
\end{table}


\section{Regularized Sparse Gaussian Processes} \label{sec:RSGP}
We propose a general regularization framework for various SGP models. This regularization framework considers not only the model fitting by the SGP objective function, but also the approximation performance of the SGP.
The approximation performance
is measured by $\int_{\bm s\in\mathbb{R}^N} |\ell_0(\bm s;\bm \theta) - \ell_1(\bm s;\bm \theta)| d\mu(\bm s)$, which is independent of the data. The smaller its value, the better the approximation performance is for that SGP model. Therefore, we naturally propose the objective function as
\begin{equation}
    \ell_2({\bm y};\bm\theta) \stackrel{\triangle}{=} \ell_1(\bm y;\bm\theta) + \lambda \int_{\bm s\in\mathbb{R}^N} |\ell_0(\bm s; \bm\theta) - \ell_1(\bm s;\bm\theta)| d\mu(\bm s)\,. \label{eq:reg2}
\end{equation}
Here, $\lambda$ is a regularization weight and $\mu(\bm s)$ is any probability measure on $\mathbb{R}^N$. For large datasets, however, computing the true log likelihood $\ell_0$ and conducting the integration with respect to all possible data are both intractable.

To avoid this problem, instead of directly measuring the difference between function $\ell_0$ and function $\ell_1$, we measure the similarity between covariance matrices $\bm Q$ and $\bm K_{nn}$ mentioned in Section~\ref{sec:background}. For most of the SGP models, the more similar the two matrices are, the better approximation the SGP achieves. In particular, for models SoR/DTC, FITC and SGPR, when $\bm Q = \bm K_{nn}$, the approximation measure $\int_{\bm s\in\mathbb{R}^N} |\ell_0(\bm s) - \ell_1(\bm s)|d\mu(\bm s)$, achieves the minimum $0$. This $\bm Q$ is called a Nystr\"om low-rank approximation of the kernel matrix $\bm K_{nn}$ in \cite{williams2000using,fowlkes2004spectral}. The error analysis of the Nystr\"om method is studied in \cite{zhang2008improved} and the  approximation is measured by $\mathcal{E} = \|{\bm K_{nn} - Q}\|_F$, where $\|\cdot\|_F$ denotes the matrix Frobenious norm. Thus we employ the same error measure to replace the regularization term in (\ref{eq:reg2}).
Alternatively, \cite{zhang2008improved} show the error of the complete kernel matrix has an upper bound that is influenced by the quantization error $\sum_{i=1}^N \|{\bm x}_i - {\bm z}_{c(i)}\|^2$, where $c(i)$ is the function that maps each sample ${\bm x}_i$ the closest inducing input ${\bm z}_{c(i)}$. If this quantization error is zero, the Nystr\"om low-rank approximation would become exact. Hence, an alternative relaxation is proposed by replacing the regularization term in (\ref{eq:reg2}) by the quantization error. We note that this quantization
is the objective function in the $k$-means clustering \cite{gersho2012vector}. 
and minimizing it is known to be an NP-hard problem \cite{drineas2004clustering}. Therefore, the same local optimum issue exists for the optimization of the proposed relaxation. Moreover, this optimization involves both continuous and discrete optimization, making the inference complicated. 

To address the optimization issue, we propose another relaxation that extends the problem to a continuous domain where gradient-based optimization methods can be directly utilized. Specifically, we assume that both data inputs and inducing inputs are independent and identically sampled from two distributions that belong to the same distribution family, i.e., ${\bm x}_i \stackrel{iid}{\sim} q_x$ and ${\bm z}_i \stackrel{iid}{\sim} q_z$, and then we encourage our objective function to decrease the distance between the two distributions. Letting $D$ be any similarity measure, we define our objective function as
\begin{eqnarray}
\ell_3 \stackrel{\triangle}{=} \ell_1 - \lambda D(\hat{q}_x, \hat{q}_z)\,, \label{eq:reg3}
\end{eqnarray}
where $\hat{q}_x = \hat{q}_x({\bm X} )$ and $\hat{q}_z = \hat{q}_z({\bm Z} )$ are estimates of $q_x$ and $q_z$, respectively, that depend on the data inputs $\bm X$ and inducing inputs $\bm Z$. 
To promote the efficiency of inference, we suggest employing a convex divergence for $D$ and the exponential family for $q$. The reason is discussed in the regularization analysis section. In this work, we employ the Kullback-Leibler (KL) divergence to measure the similarity between $\hat{q}_x$ and $\hat{q}_z$, i.e., $D(\hat{q}_x, \hat{q}_z) = \mathrm{KL}(\hat{q}_x||\hat{q}_z)$ or $D(\hat{q}_x, \hat{q}_z) = \mathrm{KL}(\hat{q}_z||\hat{q}_x)$. Furthermore we use a Gaussian distribution for $q$ and take  $\hat{q}$ to be the Gaussian with sample mean and sample covariance.
The regularized versions of the SGP models in Table~\ref{tab: SGP_objective function} are denoted by adding a prefix R to the model name. 

\subsection*{Regularization Analysis}
We claim that the objective function $\ell_3$ in (\ref{eq:reg3}) makes it easier to avoid inferior local maxima because the regularization term will guide the optimization to the manifold in which the SGP more closely approximates the full GP model.
The following theorem guarantees the convexity of the regularization  term (proof in Appendix B of the supplementary materials).

\begin{theorem} \label{theorem:convex}
Assume $q_x$ and $q_z$ belong to the same exponential family distribution. 
Then the Kullback-Leibler divergence, $KL(q_x\|q_z)$, is convex with respect to the natural parameters of $q_z$.
\end{theorem}

Due to the convexity property in Theorem~\ref{theorem:convex}, it is easier to optimize the natural parameters $\bm \theta_z$ of $q_z$. In our case, this implies the optimal $\bm Z$ are easily found  by considering the optimization of the natural parameters of $\hat{q}_z$, which are functions of $\bm Z$. This guarantees that $\bm Z$ will be close enough to a manifold where our model can achieve good approximation performance.
Therefore, when $\lambda$ is large enough, the optimization would focus more on the regularization term in (\ref{eq:reg3}) and guarantee reasonable positions for the inducing inputs. Moreover, by adjusting the regularization weight, we can improve the prediction performance.

\section{Comparative Behavior} \label{sec:CB}
The main test case in this section is the one dimensional dataset generated from this system:
\begin{eqnarray}
y|f & \sim & \mathcal{N}(y|f, 0.1^2)\,, \nonumber \\
f & = & \sin(2x) + 0.2\cos(22x) \,. \label{eq:syn_system}
\end{eqnarray} 
$100$ observations $\bm y$ and $100$ validation data $\bm y_{\text{val}}$ are generated according to (\ref{eq:syn_system}), with inputs $\bm x$ and validation inputs $\bm x_{\text{val}}$ uniformly sampled in the unit interval $[0, 1]$. We generate another $100$ evenly spaced inputs as testing inputs ${\bm x}_{test}$ on $[0,1]$ with corresponding outputs ${\bm f}_{test}$ as their ground-truth for testing. 

\subsection{Model Setting} \label{sec:model_setting}
We employ a Matern kernel with $\nu = \frac{3}{2}$ for the covariance function and set the number of inducing inputs $M = 10$. In our proposed regularized Sparse Gaussian Process models, we investigate DTC, FITC and SGPR in Section \ref{sec:tradeoff_rec_app}, \ref{sec:para_analysis} and \ref{sec:prediction_analysis} due to the direct relation between the inference and the Nystr\"om low-rank approximation $\bm Q$. In addition, we study the SVGP in Section \ref{sec:prediction_analysis}. We consider the regularization framework (\ref{eq:reg3}) with a Kullback-Leibler divergence $D(\hat{q}_x, \hat{q}_z) = \mathrm{KL}(\hat{q}_x || \hat{q}_z)$. 
The regularization weight $\lambda$ is selected from $20$ logarithmically evenly-spaced samples on $[10^{-2}, 10^2]$ by taking the optimal $\lambda_i$ with the smallest root mean square error on the validation set.  

\subsection{Trade-off between Reconstruction Performance and Approximation Performance} \label{sec:tradeoff_rec_app}
As (\ref{eq:reg3}) displays, we refer to $\ell_1$ as the reconstruction term and denote $D(\hat{q}_x, \hat{q}_z)$ to be the approximation term. The reconstruction term describes how well the model fits the data, with larger values for better fits. The approximation term shows how well the sparse Gaussian process model approximates the full Gaussian process, with smaller values indicating better performance. The balance between the reconstruction performance and the approximation performance is controlled by the regularization weight $\lambda$. We illustrate this trade-off via empirical experiments. We employ the root mean square error on the training data to summarize the reconstruction performance and borrow the approximation error measure $\mathcal{E}$ in (\ref{eq:reg3}) to represent the approximation performance. 

We investigate how our regularization framework affects the balance in DTC, FITC and SGPR in Table~\ref{tab:regularization_comparison}. Here we note that training data $\bm y$ have noise while the testing data $\bm f_{test}$ have no noise. It illustrates that our regularization always improves the approximation performance. Because of the trade-off, our regularization increases the RMSE for training data for DTC and SGPR models. For the FITC model, our regularization improve both fitting and approximation performance. This could be because our objective function helps the optimization avoid inferior local maximums as explained in the regularization analysis section. We also provide the predictive posterior process for each model in Appendix C.

\begin{table*}[ht!]
    \centering
    \caption{Model measurements on deterministic training conditional approximation (DTC), fully  independent  training  conditional  approximation (FITC) and sparse Gaussian process regression (SGPR) and their corresponding regularized models (R+model's name) as well as the full Gaussian process model (GPR). In those experiments, training data $\bm y$ contain noise while the testing data $\bm f_{test}$ have no noise.}
    \label{tab:regularization_comparison}
    \resizebox{0.8\linewidth}{!}{
    \begin{tabular}{c|c|c|c|c|c|c}
    \hline
    \multirow{2}{*}{Model} 
         & \multicolumn{2}{c|}{Trade-off measure} & Prediction measure & \multicolumn{3}{c}{Parameter} \\ 
         \cline{2-7}
        & $\text{RMSE}_{\text{train}}$ & $\mathcal{E}$ & $\text{RMSE}_{\text{test}}$ & $\beta$ & $\sigma$ & $l$ \\
        \hline
        GPR & 0.087 & - & 0.045 & 107.093 & 0.659 & 0.241 \\
        \hline
        DTC & \textbf{0.088} & 4.750 & 0.049 & \textbf{117.609} & \textbf{0.752} & \textbf{0.132} \\
        \hline
        RDTC &  0.099 & \textbf{4.411} & \textbf{0.045} & 85.218 & 0.532 & 0.124 \\
        \hline
        FITC & 0.115 & 1.822 & 0.084 & \textbf{470.545} & 0.379 & 0.154 \\
        \hline
        RFITC & \textbf{0.099} & \textbf{0.635} & \textbf{0.047} & 660.136 & \textbf{0.473} & \textbf{0.195} \\
        \hline
        SGPR & \textbf{0.098} & 0.061 & 0.044 & 79.228 & 0.781 & 0.397 \\
        \hline
        RSGPR & 0.099 & \textbf{0.049} & \textbf{0.042} & \textbf{81.353} &\textbf{0.586} & \textbf{0.311} \\
        \hline
    \end{tabular}
    }
\end{table*}

\subsection{Parameter Bias Analysis} \label{sec:para_analysis}
The estimates of the parameters for different models are shown in Table~\ref{tab:regularization_comparison}. Unlike the other cases, our regularization does not encourage the estimates of parameters for DTC to be closer to those of the full GP, even if the regularization can improve the approximation performance. This may be because DTC is a poor approximation approach for the full GP (large $\mathcal{E}$) and that imposes a barrier to learning hyper-parameters correctly. Considering the pair of FITC and RFITC approaches, except for the precision parameter of noise, our regularization encourages all parameters to be closer to the estimates of the full GP because our regularization can improve the approximation performance. The reason why it cannot learn the precision parameter of the noise well is that the diagonal correction term in FITC removes the need for heteroscedastic noise and severely underestimates the noise variance, which is thoughtfully explained in \cite{bauer2016understanding}. On the other hand, since SGPR has the best approximation performance compared with DTC and FITC, our regularization can easily encourage the values of all parameters to be closer to those of the full GP. 
\subsection{Prediction Performance Analysis} \label{sec:prediction_analysis}
We employ the root mean square of error $\text{RMSE}_{\text{test}}$ on the testing data $\bm f_{\text{test}}$ to represent the prediction performance shown in Table~\ref{tab:regularization_comparison}. Due to the consideration of the balance between the reconstruction performance and the approximation performance, our regularization improves the prediction performance for all models. 

To illustrate the robustness of our regularization framework, instead of inference on the single dataset, we repeatedly simulate $10$ independent datasets and conduct prediction analysis on them. Moreover, we explore the regularization on SVGP, which is treated as a relaxation of SGPR. The advanced prediction analysis consists of four different schedules with respect to the placement of inducing inputs. The first schedule $\mathcal{S}_1$ fixes inducing inputs as evenly spaced inputs on [0,1] and optimizes hyper-parameters through maximizing $\ell_1$. Optimizing inducing inputs during the optimization is considered as the section schedule $\mathcal{S}_2$ \cite{Snelson_2006,Titsias_2009}. Finally, we investigate our regularized schedules via (\ref{eq:reg3}) with two different similarity measures, i.e. $D(\hat{q}_x, \hat{q}_z) = \mathrm{KL}(\hat{q}_x, \hat{q}_z)$ $(\mathcal{S}_3)$ and $D(\hat{q}_x, \hat{q}_z) = \mathrm{KL}(\hat{q}_z, \hat{q}_x)$ $(\mathcal{S}_4)$. We evaluate the prediction performance using the RMSE on the testing data and show that in Table~\ref{tab:regularization_comparison_multiple}. It demonstrates that our regularization framework provides significant improvements on the model prediction performance for the four state-of-the-art sparse Gaussian process models. Additionally, we find that the order in the Kullback-Leibler divergence does not significantly affect the model prediction performance. 

\begin{table}[ht!]
	\centering
	\caption{Predictive root mean square error (RMSE) under different models for ten synthetic datasets. We summarize RMSEs' mean and standard deviation across the all datasets. The RMSEs for the full Gaussian process regression model have the mean and standard deviation 0.035(0.007).}
    \label{tab:regularization_comparison_multiple}
    \resizebox{\linewidth}{!}{
    	\begin{tabular}{c|c|c|c|c}
    		\hline
    		Model & $\mathcal{S}_1$ & $\mathcal{S}_2$ & $\mathcal{S}_3$ & $\mathcal{S}_4$ \\
    		\hline
    		DTC & 0.048(0.006) & 0.038(0.007) & \textbf{0.034(0.005)} & \textbf{0.035(0.007)}\\
    		\hline
    		FITC & 0.048(0.006) & 0.038(0.007) & \textbf{0.033(0.005)} & \textbf{0.033(0.005)} \\
    		\hline
    		SGPR & 0.052(0.008) & 0.047(0.023) & \textbf{0.035(0.009)} & \textbf{0.035(0.008)} \\
    		\hline
    		SVGP & 0.052(0.008) & 0.051(0.030) & \textbf{0.035(0.008)} & \textbf{0.035(0.008)} \\
    		\hline
    	\end{tabular}
    	}
\end{table}

\section{Regularized Latent Sparse Gaussian Processes} \label{sec:RLSGP}
The Gaussian process latent variable model (GPLVM) is a powerful dimensionality reduction approach \cite{Lawrence_2003} and it is a base model for many sophisticated models \cite{Lawrence_2007_HGP,Damianou_2016}. 
Due to the lack of scalability of GPLVM, \cite{Gal_2015} introduces inducing inputs in GPLVM to reduce the computational burden. However, it has intrinsic barriers to learning due to getting stuck in local minima, suggesting the model fitting performance is sensitive to initialization for both inducing inputs and embedding inputs. Even though principal component analysis (PCA) initialization for embedding inputs and K-means initialization for inducing inputs are standard pre-processing procedures, learning this model is problematic.

Motivated from the proposed regularization framework in SGP, we extend it to Gaussian process latent variable models. Those models treat the data inputs in SGP as random variables rather than observations. Our regularization framework is proposed for two state-of-the-art scalable latent variable models that are based on the standard GPLVM. The standard GPLVM is modelling $N$ multivariate data with dimension $D$, i.e., ${\bm Y} = \{ y_{id}\}_{i=1,d=1}^{N,D}$ with $N$ unknown embedding inputs with dimension $Q$, ${\bm X} = \{x_{iq}\}_{i=1,q=1}^{N,Q}$, where each entry $y_{id}$ is a noisy observation of the function $f_d(\bm x_i)$. All latent functions $\{f_d\}_{d=1}^D$ are given a Gaussian process prior and we let matrix $\bm F$ include $D$ vectors $\{{\bm f}_d\}_{d=1}^D$ that contain the values of the latent functions at the embedding inputs $\bm X$. We introduce a set of inducing variables ${\bm U} = \{{\bm u}_d\}_{d=1}^D$. Each of them is evaluated as ${\bm u}_d = f_d(\bm Z)$ at a set of $M$ inducing inputs $\bm Z$ with dimension $Q$. We consider a standard Gaussian prior on the embedding inputs and then we can write 

\resizebox{\linewidth}{!}{
\begin{minipage}{\linewidth}
\begin{align}
p({\bm Y}| {\bm F}) & = \prod_{d=1}^D \mathcal{N}({\bm y}_d| {\bm f}_d, \beta^{-1}{\bm I})\,, \nonumber \\
p({\bm F}| {\bm U}, {\bm X}) & = \prod_{d=1}^D \mathcal{N}({\bm f}_d| {\bm K_{nm}}{\bm K_{mm}}^{-1}{\bm u}_d, {\bm Q})\,, \nonumber \\
p({\bm U}) & = \prod_{d=1}^D\mathcal{N}({\bm u}_d| {\bm 0}, {\bm K_{mm}}) \,, \quad p({\bm X}) = \prod_{n = 1}^N\mathcal{N}({\bm x}_n| \bm\mu_X, \Sigma_X) \,, \nonumber
\end{align}
\end{minipage}
}
where $\bm K_{mm}$ is the covariance function evaluated between all the inducing inputs and $\bm K_{nm}$ is the covariance function evaluated between all the embedding inputs and inducing inputs. We define $\bm Q = \bm K_{nm}\bm K_{mm}^{-1}\bm K_{mn}$. In practice, we set prior of embedding inputs with $\bm\mu_X = \bm 0$ and $\Sigma_X = \bm I$.

Two scalable latent variables models are proposed in which \cite{Titsias_2010} marginalize the optimal distribution of inducing variables while \cite{Hensman_2012} explicitly assume the variational distribution of inducing variables belongs to the Gaussian distribution and explicitly learn it via maximizing the evidence lower bound of the log marginal likelihood. We denote these models as LSGPR and LSVGP and we denote both these models as latent sparse Gaussian process models (LSGP) and those models under our regularization framework as regularized latent sparse Gaussian process models (RLSGP).

For both LSGPR and LSVGP, variational inference can easily get trapped in inferior local maxima, which affects the model reconstruction and prediction performance. We claim that 
this behavior can be avoided
by exploiting information in the embedding inputs. In other words, obtaining a better approximation performance for the SGP associated with the latent functions $\{f_d\}_{d=1}^D$ will mitigate the chance of getting trapped in poor local maxima and deliver a better prediction result. To encourage better approximation performance, we define a new objective function. Given an evidence lower bound of the log marginal likelihood (ELBO), we introduce the modified evidence lower bound (MELBO) for either LSGPR or LSVGP. Similar to (\ref{eq:reg3}), we assume all embedding inputs $\bm X$ and all inducing inputs $\bm Z$ come from two distributions $q_x$ and $q_z$. Then we utilize the similarity between $\hat{q}_x$ and $\hat{q}_z$ to quantify the approximation performance of the SGP for the latent functions, where $\hat{q}_x$ and $\hat{q}_z$ are the estimates of $q_x$ and $q_z$, respectively. The MELBO can then be expressed as
\begin{eqnarray}
    \mathrm{MELBO} = \mathrm{ELBO} - \lambda D(\hat{q}_x, \hat{q}_z)\,. \label{eq:reg_latent}
\end{eqnarray}

Since in the latent variable models the embedding variables $\bm X$ are unknown with Gaussian priors, we must change our assumptions of $q_x$. Suppose the variational distribution of embedding inputs is $q(\bm X) = \prod_{n=1}^N\mathcal{N}({\bm x}_n| {\bm \mu}_n, \Sigma_n)$. We assume the variational mean of $q(\bm X)$, i.e., $\{{\bm \mu}_n\}_{n=1}^N$ are sampled from $q_x$, and we estimate $q_x$ with those variational means. As for $q_z$, we estimate it with the inducing inputs $\bm Z$ as before.

As with the SGP case, the regularization weight plays an important role in the balance between the reconstruction performance and approximation performance. Although there is a trade-off between reconstruction and approximation performance, with the proper choice of regularization weight, our model can improve both in practice due to the convexity of the regularization term as discussed previously. 

\subsection*{Relation between Regularization Framework and Empirical Bayesian Model}

In this section we present a theorem that demonstrates a connection between our proposed regularization method and an empirical Bayes method. 
Specifically, we show that variational inference for the regularized LSVG model is equivalent to maximizing a variational lower bound in a corresponding empirical Bayes model. 

The empirical Bayes model is formulated with a prior specification over the inducing inputs $Z$. After deriving a variational lower bound with a structured variational distribution, we show that when $\lambda = M$ and $D(\hat{q}_x, \hat{q}_z) = \mathrm{KL}(\hat{q}_z||\hat{q}_x)$, maximizing the MELBO is equivalent to maximizing this variational lower bound in the empirical Bayesian model under a mild condition. Specifically, the empirical Bayesian model is extended from the LSVGP by considering an informative prior $p(\bm Z)$ with its variational distribution $q(\bm Z)$ on inducing inputs as 
\begin{eqnarray}
p(\bm z_m) & = & \mathcal{N}(\bm z_m| \hat{\bm \mu}_{\bm\mu}, \hat{\Sigma}_{\bm\mu})\,, \nonumber \\
q(\bm z_m) & = & \mathcal{N}(\bm z_m|\bm \nu_m, \Upsilon = \epsilon \bm I)\,,  \nonumber
\end{eqnarray}
where $\hat{\bm \mu}_{\bm \mu}$, $\hat{\Sigma}_{\bm \mu}$ are estimates using the sample mean and sample covariance matrix of $\{\bm \mu_n\}$. 

\begin{figure}[ht!]
	\centering
	\includegraphics[width=\linewidth]{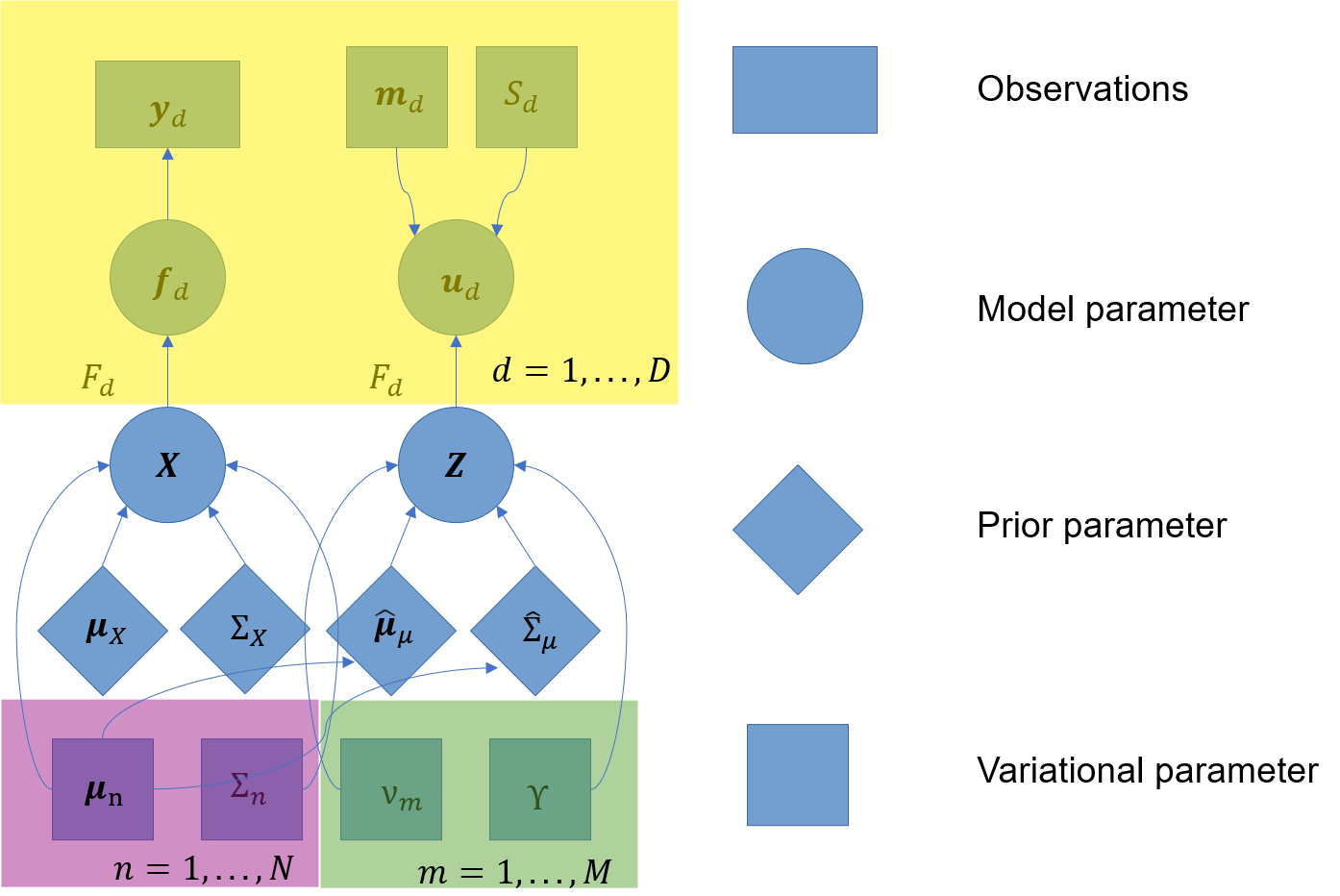}
	\caption{Graphical representation for the empirical Bayesian model. The yellow, purple and green regions show the independent modeling with respect to output dimension $d$, individual data $n$ and individual inducing data $m$.}
	\label{fig:EB}
\end{figure}

In the LSVG model, we assume the varaitional distribution is $q(\bm u_d) = \mathcal{N}(\bm u_d| \bm m_d, S_d)$. The empirical Bayesian model is displayed using a graphical representation in Figure~\ref{fig:EB}. The prior for the inducing points borrows the information from the variational mean of the embedding inputs $\bm \mu$. The variational joint distribution is proposed as 
\begin{align}
    q(\bm F, \bm U, \bm X, \bm Z) = q(\bm Z)q(\bm X)q(\bm U)p(\bm F|\bm Z, \bm X, \bm U)\,. \nonumber
\end{align}

Letting $\hat{\bm \mu}_{\bm\nu}$ and $\hat{\Sigma}_{\bm\nu}$ be the sample mean and sample covariance matrix of $\{\bm \nu_m\}$, we assume there exist values $K_1$ and $K_2$ such that $0 < K_1 < |\hat{\Sigma}_{\bm \nu}| < K_2$. Given this mild condition, we prove that variational inference for the regularized LSVG model is equivalent to maximizing a variational lower bound for this empirical Bayes model in the following theorems. 

\begin{theorem} \label{theorem:lower_bound}
For any $\epsilon > 0$, we have the lower bound of log marginal likelihood such that
\begin{align}
    \log p(\bm Y) & \geq  \mathrm{E}_{q(\bm F, \bm U, \bm X, \bm Z)}\log p(\bm Y|\bm F) - \mathrm{KL}(q(\bm Z)||p(\bm Z)) \nonumber \\
    & - \mathrm{KL}(q(\bm X)||p(\bm X)) - \mathrm{KL}(q(\bm U)||p(\bm U)) \nonumber \\
    & \stackrel{\triangle}{=} \mathrm{ELBO}_{\mathrm{EB}}(\epsilon)\nonumber \\
    & \geq  \mathrm{E}_{q(\bm F, \bm U, \bm X, \bm Z)}\log p(\bm Y|\bm F) - A + B + C \nonumber \\
    & - \mathrm{KL}(q(\bm X)||p(\bm X)) - \mathrm{KL}(q(\bm U)||p(\bm U)) \nonumber \\
    & \stackrel{\triangle}{=} \mathrm{LELBO}_{\mathrm{EB}}(\epsilon)\,.
\end{align}
where
\begin{align}
	& A = \frac{M}{2}(\log |\hat{\Sigma}_{\bm \mu}|  + \log|\hat{\Sigma}_{\bm\nu}| + Q) \nonumber \\
	& \quad + \frac{1}{2}\left(\sum_{m = 1}^M(\bm\nu_m - \hat{\bm \mu}_{\bm \mu})^T\hat{\Sigma}_{\bm\mu}^{-1}(\bm\nu_m - \hat{\bm \mu}_{\bm \mu})\right), \nonumber \\
	& B = \frac{M}{2}(Q\log\epsilon-\log K_2), \nonumber \\
	& C = \frac{2\epsilon}{M\mathrm{tr}(\hat{\Sigma}_{\bm\mu}^{-1})} \nonumber \,.
\end{align}
Moreover, the difference between two lower bounds is bounded by
\begin{align}
    \mathrm{ELBO}_{\mathrm{EB}} - \mathrm{LELBO}_{\mathrm{EB}} \leq \frac{M}{2}(\log K_1 - \log K_2)\,.
\end{align}
\end{theorem}


\begin{theorem} \label{theorem:relation}
As $\epsilon \downarrow 0$, maximizing the lower bound $\mathrm{LELBO}_{\mathrm{EB}}$ is equivalent to maximizing $\mathrm{MELBO}$ in which $\lambda=M$ and $D(\hat{q}_x, \hat{q}_z) = \mathrm{KL}(\hat{q}_z||\hat{q}_x)$.
\end{theorem}

The proofs of Theorem~\ref{theorem:lower_bound} and Theorem~\ref{theorem:relation} are provided in Appendix D of the supplementary material. 

\section{Experiments} \label{sec:experiments}
We illustrate our regularization framework on two real datasets. First, we show that regularization contributes to better model fitting in latent variable models on the Anuran Calls data. In this experiment, we explore the regularization for both LSGPR and LSVGP and exploit how different latent dimension sizes affect the model performance in the regularization framework. Second, we demonstrate our regularization framework on a large dataset with different numbers of inducing points using the Flight dataset \cite{Hensman_2013}. All optimizations employ the limited memory Broyden–Fletcher–Goldfarb–Shanno with boundaries (L-BFGS-B) method \cite{zhu1997algorithm} with maximum iteration number $1000$. We employ the squared exponential covariance function with an automatic relevance determination (ARD) structure \cite{zhao2018variational} in both models for all experiments. All experiments are run on Ubuntu system with Intel(R) Core(TM) i7-7820X CPU @ 3.60GHz and 128G memory. 

\subsection{Anuran Calls Example} \label{sec:Anuran}
We show that regularization improves inference on the Anuran Call dataset. This dataset is available from the UCI repository at \url{https://archive.ics.uci.edu/ml/datasets/Anuran+Calls+(MFCCs)}, where there are $7195$ instances, and each instance has $22$ attributes. We model all instances using a sparse latent Gaussian process and perform inference with and without regularization. Moreover, to account for variability in reconstruction performance, we consider 10 different subsets. Each subset includes $4000$ instances, randomly sampled from the full dataset without replacement.

We set the latent dimension size $Q = 5$ and use $M = 20$ inducing points in RLSGP models. We employ the PCA approach for initialization of the embedding inputs and the K-means algorithm for initialization of the inducing inputs, with all other hyper-parameters initialized according to \cite{Titsias_2010} and \cite{Hensman_2013}.

We conduct experiments on both LSGPR, LSVGP and their regularized models under three schedules. The first schedule $\mathcal{S}_1$ fixes the inducing inputs as the initial K-means' centroids. The second schedule $\mathcal{S}_2$ treats the inducing inputs as trainable parameters in optimization. The last schedule $\mathcal{S}_3$ considers our proposed regularization approach in (\ref{eq:reg_latent}). The regularization $\lambda$ is selected from $[1, 10, 100, 1000]$ with the smallest root mean square error (RMSE) in the model reconstruction. We estimate embedding inputs as their variational mean $\hat{X} = \hat{\bm \mu}$ and reconstruct all observations by the posterior mean at embedding inputs $\hat{X}$. We compare the similarity of distributions between embedding inputs $X$ and inducing inputs $Z$ by a proposed averaged symmetric KL divergence criterion (ASKL) defined by $\mathrm{ASKL} = \frac{1}{Q}\sum_{q=1}^Q(0.5\mathrm{KL}(\hat{p}(\hat{X}_q)||\hat{p}(\hat{Z}_q)+0.5\mathrm{KL}(\hat{p}(\hat{Z}_q)||\hat{p}(\hat{X}_q))$ where $\hat{p}(X)$ is a Gaussian distribution fitted by $X$. The ASKL measures the average similarity between $X$ and $Z$ across all latent dimensions. 

\begin{table*}[!htb]
	\centering
	\caption{Anuran Calls Example: Root mean square error (RMSE) and averaged symmetric KL divergence (ASKL) for three different schedules with respect to inducing inputs for LSGPR and LSVGP models. We consider inference on the full dataset and 10 randomly sampled subsets separately (with mean and standard deviation).}
	\label{tab:Anuran}
	\resizebox{0.8\linewidth}{!}{
	\begin{tabular}{c|c|c|c|c|c|c|c}
		\hline
		\multicolumn{2}{c|}{} & \multicolumn{3}{|c|}{Full Data} & \multicolumn{3}{|c}{Subset} \\
		\hline
		\multicolumn{2}{c|}{Model} & $S_1$ & $S_2$ & $S_3$ & $S_1$ & $S_2$ & $S_3$ \\
		\hline
		\multirow{2}{*}{LSGPR} &
		RMSE & 0.058 & 0.044 & \textbf{0.043} & 0.052(0.0019) & 0.043(0.0005) & 0.042(0.0003) \\
		\cline{2-8}
		& ASKL & 2.533 & 0.421 & \textbf{0.011} & 3.423(0.3973) & 0.077(0.0543) & 0.011(0.0019)\\
		\hline
		\multirow{2}{*}{LSVGP} &
		RMSE & 0.078 & 0.050 & \textbf{0.046} & 0.071(0.0023) & 0.048(0.0008) & 0.044(0.0005) \\
		\cline{2-8}
		& ASKL & 4.241 & 27.545 & \textbf{ 0.140} & 3.045(0.7630) & 16.898(8.1971) & 0.025(0.0051)\\
		\hline
	\end{tabular}
	}
\end{table*}

\subsubsection{Regularization with different methods}
We explore both LSGPR and LSVGP models on the full dataset and we also carry out experiments on 10 randomly sampled subsets to show the robustness of results. RMSEs and ASKLs are summarized in Table~\ref{tab:Anuran}, showing that our regularization approach contributes to better performance on both model fitting and latent input deployment. 
Also, LSVGP under $\mathcal{S}_2$ has a significantly larger ASKL compared with the same schedule using LSGPR, because without marginalization, the nonlinear, non-convex objective function for LSVGP involves more parameters making optimization difficult. However, with our proposed regularization, this model gets a comparable model fitting result.


\subsubsection{Regularization with different latent dimension sizes}

Although LSGPR has better fitting performance than LSVGP, it is not scalable and is less robust in the optimization. Here we only consider LSVGP with its regularized version and explore the benefits of regularization for different latent dimension sizes. Because of output dimension size $D=22$, we consider the different latent dimension sizes $Q = 2,5,10$ and set regularization weight $\lambda = 1000$ which performs well for this dataset. The RMSEs and ASKLs are displayed in Table~\ref{tab:Anuran_lamb}, showing the consistent benefits of regularization, especially when $Q$ is much smaller than $D$. 

\begin{table}[ht!]
	\centering
	\caption{Anuran Calls Example: Root mean square errors (RMSE) and averaged symmetric KL divergence (ASKL) for models with fixed inducing inputs (F), and models with learned inducing inputs without regularization (N) and with regularization (R) under latent dimension sizes $Q = 2, 5, 10$.}
	\label{tab:Anuran_lamb}
	\resizebox{0.9\linewidth}{!}{
	\begin{tabular}{c|c|c|c}
		\hline
		& Q = 2 & Q = 5 & Q = 10 \\
		\hline
		RMSE(F) & 0.0871 & 0.0788 & 0.0754 \\
		\hline
		RMSE(N) & 0.0851 & 0.0500 & 0.0357 \\
		\hline
		RMSE(R) & \textbf{0.0655} & \textbf{0.0487} & \textbf{0.0324} \\
		\hline
		ASKL(F) & 1.8147 & 4.1234 & 5.4049 \\ 
		\hline
		ASKL(N) & 1315.6931 & 33.7854 & 82.1224 \\
		\hline
		ASKL(R) & \textbf{0.5898} & \textbf{0.0487} & \textbf{0.0141} \\
		\hline
	\end{tabular}
	}
\end{table}

\subsection{Flight Example}
Our second example is the Flight data, which consists of every commercial flight in the USA from January to April 2008, information on 2 million flights. We include the same 8 variables as in \cite{Hensman_2013}. Instead of predicting the delay time using the 8 features, we focus on reconstruction tasks on both noiseless data and noisy data. Specifically, we first standardize data within each feature and then we randomly select 10k flights and randomly choose one of the 8 features to add white noise in those 10k flights' data. Our task is to reconstruct the features for 10k noisy flights' data and the other 70k noiseless fights' data. We compare the reconstructed features with ground truth noiseless features. We repeat the noisy data generation procedure 10 times with different random seeds to obtain 10 datasets to measure robustness of our method.

We take LSVGP as the baseline model with latent dimension size $Q = 2$ and $M = 20$ inducing points. RLSVGP is applied to the whole training dataset. Based on the reconstruction performance, we select the regularization weight $\lambda = 100$ out of a candidate pool $[1, 10, 100, 1000, 10000]$.

RMSEs for both the 70k noiseless data and 10k noisy data are displayed in Table~\ref{tab:Flight}. We summarize the mean and standard deviation of RMSEs over the ten randomly generated datasets. The result shows that our model outperforms the baseline model on both model fitting and noisy data reconstruction tasks. 


\begin{table}[ht!]
	\centering
	\caption{Flight Example: Root mean square errors of 70k noiseless data and 10k noisy data for baseline/regularized model.}
	\label{tab:Flight}
	\resizebox{\linewidth}{!}{
	\begin{tabular}{c|c|c|c|c}
		\hline
		 & Model & M = 10 & M = 20 & M = 50 \\
		\hline
		\multirow{2}{*}{Noiseless data} &
		Baseline     & 0.772(0.0049) & 0.660(0.0022) & 0.643(0.0056)  \\
		\cline{2-5}
		& RGPLVM & \textbf{0.654(0.0019)} & \textbf{0.614(0.0135)} & \textbf{0.614(0.0112)} \\
		\hline
		\multirow{2}{*}{Noisy data}
		& Baseline & 0.818(0.0043) & 0.674(0.0021) & 0.663(0.0108)  \\
		\cline{2-5}
		& RGPLVM & \textbf{0.668(0.0029)} & \textbf{0.631(0.0121)} & \textbf{0.634(0.0113)} \\
		\hline
	\end{tabular}
	}
\end{table}

\section{Conclusion}
\label{sec:conclusion}
To consider the quality of the approximation for inducing-input based sparse Gaussian process models, we proposed a novel regularization framework that balances the reconstruction and approximation performance and so improves model prediction performance. Our framework also alleviates the local maxima issue by exploiting the convexity of the regularization term. We thoroughly investigated this regularization in terms of the model's objective function and the behavior observed during practical applications. 

We then extended this framework to latent variable models and provide a relation between the regularization approach and empirical priors theoretically. Finally, we illustrate that our regularization framework robustly improves model inference and prediction.

\clearpage
\nocite{langley00}

\bibliography{main}
\bibliographystyle{icml2021}





\end{document}


\twocolumn[
\icmltitle{Supplementary Material}



\icmlsetsymbol{equal}{*}

\begin{icmlauthorlist}
\icmlauthor{Aeiau Zzzz}{equal,to}
\icmlauthor{Bauiu C.~Yyyy}{equal,to,goo}
\icmlauthor{Cieua Vvvvv}{goo}
\icmlauthor{Iaesut Saoeu}{ed}
\icmlauthor{Fiuea Rrrr}{to}
\icmlauthor{Tateu H.~Yasehe}{ed,to,goo}
\icmlauthor{Aaoeu Iasoh}{goo}
\icmlauthor{Buiui Eueu}{ed}
\icmlauthor{Aeuia Zzzz}{ed}
\icmlauthor{Bieea C.~Yyyy}{to,goo}
\icmlauthor{Teoau Xxxx}{ed}
\icmlauthor{Eee Pppp}{ed}
\end{icmlauthorlist}

\icmlaffiliation{to}{Department of Computation, University of Torontoland, Torontoland, Canada}
\icmlaffiliation{goo}{Googol ShallowMind, New London, Michigan, USA}
\icmlaffiliation{ed}{School of Computation, University of Edenborrow, Edenborrow, United Kingdom}

\icmlcorrespondingauthor{Cieua Vvvvv}{c.vvvvv@googol.com}
\icmlcorrespondingauthor{Eee Pppp}{ep@eden.co.uk}

\icmlkeywords{Machine Learning, ICML}

\vskip 0.3in
]



\printAffiliationsAndNotice{\icmlEqualContribution} 

\appendix

\section{Sparse Gaussian Process Models}

As for those sparse Gaussian Process models, the predictive posterior mean $\mu(\cdot)$ at any new input $x$ can be treated as a weighted sum of kernels centered at the inducing inputs \cite{Rasmussen_2005}. They are derived in the same form such that $\mu(x) = {\bf k}_Z(x)\bf c$, where ${\bf k}_Z(x)$ is the covariance function evaluated between all the inducing inputs and the new input and $\bf c$ is a weight vector. This weight vector is specified differently in each SGP model. We present the wight vector in the expression of predictive posterior mean in SoR/DTC, FITC, SGPR and SVGP models in Table~\ref{tab: SGP_predictive_mean}.
\begin{table}[ht!]
	\centering
	\caption{Weight vector in the expression of predictive posterior mean in different sparse Gaussian process models.}
	\label{tab: SGP_predictive_mean}
	\resizebox{0.8\linewidth}{!}{
	\begin{tabular}{c|c}
		\hline
		MODEL & Weight vector $\bf c$ \\
		\hline 
		SoR/DTC & $\bf K_{mm}^{-1}\bf K_{mn}(Q + \beta^{-1}I)^{-1}\bf y$ \\
		\hline
		FITC & $\bf K_{mm}^{-1}\bf K_{mn}(Q + \mathrm{diag}(\tilde{\bf K}) + \beta^{-1}I)^{-1}\bf y$ \\
		\hline
		SGPR & $\beta^{-1}(\bf K_{mm} + \beta\bf K_{mn}\bf K_{nm})^{-1}\bf K_{mn}\bf y$ \\
		\hline
		SVGP & $\bf K_{mm}^{-1} \bf m$ \\
		\hline
	\end{tabular}
	}
\end{table}

\section{Regularization Analysis}
Proof of Theorem 1 in the main paper. 
\begin{proof}
For an exponential family distribution 
\begin{align}
    q(\bm y| \bm \eta) \propto \exp(\langle S(\bm y), \bm \eta \rangle - A(\bm \eta))\,, \nonumber
\end{align}
where $S$, $\bm \eta$, and $A$ are the sufficient statistics, natural parameters, and log-partition function (cumulant function) of the distribution. The log-partition function is strictly convex in a minimal family \cite{rockafellar1970convex}. We would abuse notation $D(\hat{q}_x||\hat{q}_z) = D(\bm \eta_x||\bm \eta_z)$ and the KL divergence admits a closed form expression \cite{nielsen2012closed}. 
\begin{align}
    D(\bm \eta_x\|\bm \eta_z) & = \langle \bm \eta_x - \bm \eta_z, \bigtriangledown A(\bm \eta_x) \rangle - A(\bm\eta_x) + A(\bm\eta_z) \nonumber \\
    & = B_{A}(\bm \eta_z, \bm\eta_x) \nonumber
\end{align}
where $B_{A}(\bm \eta_z, \bm\eta_x)$ is a Bregman divergence computed on the swapped natural parameters.
Because $A(\bm\eta_z)$ is strictly convex and $\bm\eta_z^T\bigtriangledown A(\bm \eta_x)$ is an affine transformation, $D(\bm \eta_x\|\bm \eta_z)$ is strictly convex in $\bm \eta_z$. 
\end{proof}

\section{Predictive Posterior Process for Sparse Gaussian Process Models}
\begin{figure*}[ht!]
    \centering
    \minipage{0.32\textwidth}
        \includegraphics[trim=5mm 0mm 15mm 5mm,clip,width = \linewidth]{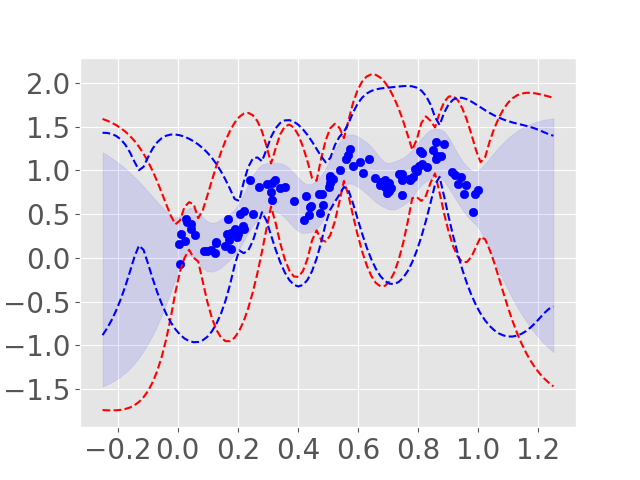}
        \caption*{(a) Predictive posterior process for DTC}
    \endminipage
    \hfill
    \minipage{0.32\textwidth}
        \includegraphics[trim=5mm 0mm 15mm 5mm,clip,width = \linewidth]{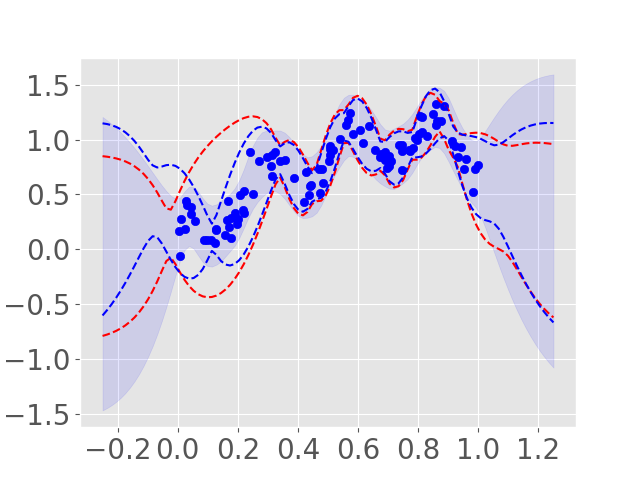}
        \caption*{(b) Predictive posterior process for FITC}
    \endminipage
    \hfill
    \minipage{0.32\linewidth}
        \includegraphics[trim=5mm 0mm 15mm 5mm,clip,width = \linewidth]{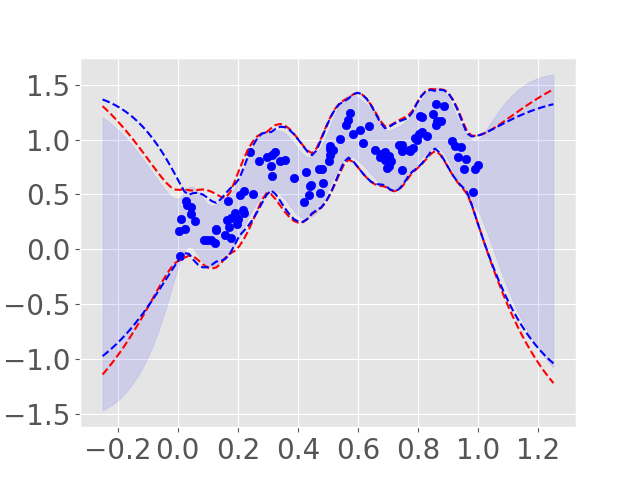}
        \caption*{(b) Predictive posterior process for SGPR}
    \endminipage
    \caption{Behavior of deterministic training conditional approximation (DTC), fully  independent  training  conditional  approximation (FITC) and sparse Gaussian process regression (SGPR) on a set of 100 data simulated from default system using 10 inducing inputs (blue points indicate data; red dashed lines indicates mean$\pm$2$\sigma$ for original method; blue dashed lines indicates mean$\pm$2$\sigma$ for regularized method) compared to the prediction of the full GP in grey.}
    \label{fig:regularization_comparison}
\end{figure*}

We plot the predictive  posterior process for three pairs of sparse Gaussian process models including DTC, FITC and SGPR in Figure~\ref{fig:regularization_comparison}. It shows that the confidence band for RFITC is significantly closer to that for the full GP compared with FITC, while the comparison for the other two pairs of models is less visually clear. 

\section{Relation between Regularization Framework and Empirical Bayesian Model}

Proof of Theorem 2 in the main paper.

\begin{proof}
    According to the Jensen's inequality, we have 
    \begin{align}
        \log p(\bm Y) & \geq  \mathrm{E}_{q(\bm F, \bm U, \bm X, \bm Z)}\log p(\bm Y|\bm F) - \mathrm{KL}(q(\bm Z)||p(\bm Z)) \nonumber \\
    & - \mathrm{KL}(q(\bm X)||p(\bm X)) - \mathrm{KL}(q(\bm U)||p(\bm U))\nonumber \\
    & \stackrel{\triangle}{=} \mathrm{ELBO}_{\mathrm{EB}} \label{eq:ELBO}
    \end{align}
    Given the definitions of $A, B, C$ in the Theorem 2, we have
    \begin{align}
        \mathrm{KL}(q(Z)|| p(Z)) & =
		A - \frac{M}{2}(Q\log\epsilon-\log |\hat{\Sigma}_{\bm \nu}|) - C\,. \nonumber 
    \end{align}
    Because $K_1 < |\hat{\Sigma}_{\bm \nu}| < K_2$, we have that 
    \begin{align}
        \frac{M}{2}(Q\log\epsilon - \log K_2) \leq \frac{M}{2}(Q\log\epsilon - \log|\hat{\Sigma}_{\bm \nu}|) \nonumber\\ 
        \quad \leq \frac{M}{2}(Q\log\epsilon - \log K_1)\,. \nonumber
    \end{align}
    It suggests that $\mathrm{KL}(q(\bm Z)||p(\bm Z))$ is bounded by
    \begin{align}
        A - B_1 - C \leq \mathrm{KL}(q(\bm Z)||p(\bm Z)) \leq A -  B_2 - C\,, \label{eq:ineq}
    \end{align}
    where $B_1 = \frac{M}{2}(Q\log\epsilon - \log K_1)$ and $B_2 = \frac{M}{2}(Q\log\epsilon - \log K_2)$.
    Plugging (\ref{eq:ineq}) into the ELBO (\ref{eq:ELBO}), we have a lower bound (\ref{eq:LELBO}) and a upper bound (\ref{eq:UELBO}) for ELBO such that
    \begin{align}
        \mathrm{ELBO}_{\mathrm{EB}} & \geq \mathrm{ELBO}_{\mathrm{EB}} - \mathrm{KL}(q(\bm Z)||p(\bm Z)) + A -  B_1 - C \nonumber \\
        & \stackrel{\triangle}{=} \mathrm{UELBO}_{\mathrm{EB}} \,, \label{eq:LELBO} \\
        \mathrm{ELBO}_{\mathrm{EB}} & \leq \mathrm{ELBO}_{\mathrm{EB}} - \mathrm{KL}(q(\bm Z)||p(\bm Z)) + A -  B_2 - C \nonumber \\
        & \stackrel{\triangle}{=} \mathrm{LELBO}_{\mathrm{EB}} \,. \label{eq:UELBO}
    \end{align} 
    Then the difference between $\mathrm{ELBO}_{\mathrm{EB}}$ and $\mathrm{LELBO}_{\mathrm{EB}}$ is bound by
    \begin{align}
        \mathrm{ELBO}_{\mathrm{EB}} - \mathrm{LELBO}_{\mathrm{EB}} & \leq \mathrm{UELBO}_{\mathrm{EB}} - \mathrm{LELBO}_{\mathrm{EB}} \nonumber \\
        & = \frac{M}{2}(\log K_1 - \log K_2)\,.
    \end{align}
\end{proof}

Before we give the proof of Theorem 3 in the main paper, we give two lemmas as follows:

\begin{lemma}
	As $\epsilon \rightarrow 0$, $\bm z_m \stackrel{p}{\rightarrow} \bm \nu_m$.
\end{lemma}

\begin{proof} 
	Because $\forall \epsilon_0 > 0$,
	\begin{eqnarray}
	\lim\limits_{\epsilon \rightarrow 0} p(|\bm z_m - \bm \nu_m| > \epsilon_0) & = & \lim\limits_{\epsilon \rightarrow 0} p(|\frac{\bm z_m - \bm \nu_m}{\epsilon}| > \frac{\epsilon_0}{\epsilon}) \nonumber \\
	& = & 2\lim\limits_{\epsilon \rightarrow 0} (1 - \Phi(\frac{\epsilon_0}{\epsilon}))^{Q} \nonumber \\
	& = & 0 \,, \nonumber
	\end{eqnarray}
	we conclude that as $\epsilon \rightarrow 0$, $\bm z_m \stackrel{p}{\rightarrow} \bm \nu_m$.
\end{proof}

\begin{lemma}
	\begin{align}
	& M\mathrm{KL}(\hat{q}_z || \hat{q}_x) = \frac{M}{2}(\log |\hat{\Sigma}_{\bm \mu}| + \log|\hat{\Sigma}_{\bm z}| + Q) \nonumber \\
	& + \frac{1}{2}\left(\sum_{m = 1}^M(\bm z_m - \hat{\bm \mu}_{\bm \mu})^T\hat{\Sigma}_{\bm\mu}^{-1}(\bm z_m - \hat{\bm \mu}_{\bm \mu})\right) \nonumber\,.
	\end{align}
\end{lemma}

\begin{proof}
	{\footnotesize
		\begin{align}
		& \mathrm{KL}(\hat{q}_z || \hat{q}_x) \nonumber \\
		& =  \frac{1}{2}\left[\log\frac{|\hat{\Sigma}_{\bm\mu}|}{|\hat{\Sigma}_z|} - Q + \mathrm{tr}(\hat{\Sigma}_{\bm\mu}^{-1}\hat{\Sigma}_z)+(\hat{\bm \mu}_{\bm\mu} - \hat{\bm \mu}_z)^T\hat{\Sigma}_{\bm\mu}^{-1}(\hat{\bm \mu}_{\bm\mu} - \hat{\bm \mu}_z)\right] \nonumber \\
		& =  \frac{1}{2}\left[\log\frac{|\hat{\Sigma}_{\bm\mu}|}{|\hat{\Sigma}_z|} - Q + \mathrm{tr}\left(\hat{\Sigma}_{\bm\mu}^{-1}((\hat{\bm \mu}_{\bm\mu} - \hat{\bm \mu}_z)(\hat{\bm \mu}_{\bm\mu} - \hat{\bm \mu}_z)^T + \hat{\Sigma}_z)\right)\right] \nonumber \\
		& =  \frac{1}{2}\Bigg[\log\frac{|\hat{\Sigma}_{\bm\mu}|}{|\hat{\Sigma}_z|} - Q + \frac{1}{M}\mathrm{tr}\bigg(\hat{\Sigma}_{\bm\mu}^{-1}(M(\hat{\bm \mu}_{\bm\mu} - \hat{\bm \mu}_z)(\hat{\bm \mu}_{\bm\mu} - \hat{\bm \mu}_z)^T + \nonumber \\
		& \quad \sum_{m = 1}^{M}(\bm u_m - \hat{\bm \mu}_z)(\bm u_m - \hat{\bm \mu}_z)^T\bigg)\Bigg] \nonumber \\
		& =  \frac{1}{2}\Bigg[\log\frac{|\hat{\Sigma}_{\bm\mu}|}{|\hat{\Sigma}_Z|} - Q + \frac{1}{M}\mathrm{tr}\bigg(\hat{\Sigma}_{\bm\mu}^{-1}(M\hat{\bm \mu}_{\bm\mu}\hat{\bm \mu}_{\bm\mu}^T - M\hat{\bm \mu}_{\bm\mu}\hat{\bm \mu}_z^T - M\hat{\bm \mu}_z\hat{\bm \mu}_{\bm\mu}^T + \nonumber \\
		& \quad M\hat{\bm \mu}_z\hat{\bm \mu}_z^T + \sum_{m = 1}^{M}\bm u_m \bm u_m^T - (\sum_{m = 1}^{M}\bm z_m)\hat{\bm \mu}_z^T - \hat{\bm \mu}_Z(\sum_{m = 1}^{M}\bm z_m)^T + M\hat{\bm \mu}_z\hat{\bm \mu}_z^T \bigg)\Bigg] \nonumber \\
		& =  \frac{1}{2}\Bigg[\log\frac{|\hat{\Sigma}_{\bm\mu}|}{|\hat{\Sigma}_z|} - Q + \frac{1}{M}\mathrm{tr}\bigg(\hat{\Sigma}_{\bm\mu}^{-1}(\sum_{m = 1}^{M}\bm z_m\bm z_m^T - M\hat{\bm \mu}_{\bm\mu}\hat{\bm \mu}_Z^T - M\hat{\bm \mu}_Z\hat{\bm \mu}_{\bm\mu}^T + \nonumber \\
		& \quad M\hat{\bm \mu}_X\hat{\bm \mu}_{\bm\mu}^T)\bigg)\Bigg]\nonumber \\
		& =  \frac{1}{2}\Bigg[\log\frac{|\hat{\Sigma}_{\bm\mu}|}{|\hat{\Sigma}_z|} - Q + \frac{1}{M}\mathrm{tr}\bigg(\hat{\Sigma}_{\bm\mu}^{-1}(\sum_{m = 1}^{M}\bm z_m\bm z_m^T - \hat{\bm \mu}_{\bm\mu}(\sum_{m = 1}^{M}\bm z_m)^T - \nonumber \\
		& \quad (\sum_{m = 1}^{M}\bm z_m)\hat{\bm \mu}_{\bm\mu}^T + M\hat{\bm \mu}_{\bm\mu}\hat{\bm \mu}_{\bm\mu}^T)\bigg)\Bigg]\nonumber \\
		& =  \frac{1}{2}\Bigg[\log\frac{|\hat{\Sigma}_{\bm\mu}|}{|\hat{\Sigma}_z|} - Q + \frac{1}{M}\mathrm{tr}\bigg(\hat{\Sigma}_{\bm\mu}^{-1}(\sum_{m = 1}^{M}\bm z_m\bm z_m^T - \hat{\bm \mu}_{\bm\mu}(\sum_{m = 1}^{M}\bm u_m)^T - \nonumber \\
		& \quad (\sum_{m = 1}^{M}\bm z_m)\hat{\bm \mu}_{\bm\mu}^T + M\hat{\bm \mu}_{\bm\mu}\hat{\bm \mu}_{\bm\mu}^T)\bigg)\Bigg]\nonumber \\
		& =  \frac{1}{2}\Bigg[\log\frac{|\hat{\Sigma}_{\bm\mu}|}{|\hat{\Sigma}_z|} - Q + \frac{1}{M}\mathrm{tr}\bigg(\hat{\Sigma}_{\bm\mu}^{-1}(\sum_{m = 1}^{M}(\bm z_m - \hat{\bm \mu}_{\bm\mu})(\bm z_m - \hat{\bm \mu}_{\bm\mu})^T)\bigg)\Bigg]\nonumber \\
		& =  \frac{1}{M}\sum_{m = 1}^{M}\frac{1}{2}\left[\log\frac{|\hat{\Sigma}_{\bm\mu}|}{|\hat{\Sigma}_z|} - Q + (\bm z_m - \hat{\bm \mu}_{\bm\mu})^T\hat{\Sigma}_{\bm\mu}^{-1}(\bm z_m - \hat{\bm \mu}_{\bm\mu})\right]\nonumber \,.
		\end{align}
	}
	Therefore, 
	{\footnotesize
	\begin{align}
	& M\mathrm{KL}(\hat{q}_z || \hat{q}_x) \nonumber\\
	& = \frac{1}{2}\sum_{m = 1}^{M}\left[\log\frac{|\hat{\Sigma}_{\bm\mu}|}{|\hat{\Sigma}_z|} - Q + (\bm z_m - \hat{\bm \mu}_{\bm\mu})^T\hat{\Sigma}_{\bm\mu}^{-1}(\bm z_m - \hat{\bm \mu}_{\bm\mu})\right] \nonumber \\
	& = \frac{M}{2}\left(\log |\hat{\Sigma}_{\bm \mu}| + \log|\hat{\Sigma}_{z}| + Q\right) + \nonumber \\
	& \quad \frac{1}{2}\left(\sum_{m = 1}^M(\bm z_m - \hat{\bm \mu}_{\bm \mu})^T\hat{\Sigma}_{\bm\mu}^{-1}(\bm z_m - \hat{\bm \mu}_{\bm \mu})\right)\,. \nonumber
	\end{align}
	}
	
\end{proof}

Proof of Theorem 2 in the main paper.

\begin{proof}
	{\footnotesize
		In the empirical Bayesian model, we denote all parameters as $\varTheta = [\bm \mu, \bm \Sigma, \bm m, \bm S, \bm \nu, \bm h]$ where $\bm h$ denote all hyper-parameters in GP kernels. 
		
		First of all, according to Lemma 1, 
		\begin{align}
		\lim\limits_{\epsilon \rightarrow 0}E_{q(F, U, X, Z)}\log p(Y|F) = E_{q(F, U, X)}\log p(Y|F, Z = \bm \nu)\,. \nonumber
		\end{align}
		Second, $\lim\limits_{\epsilon \rightarrow 0}C =  \frac{2}{M\mathrm{tr}(\hat{\Sigma}_{\mu}^{-1})}\lim\limits_{\epsilon \rightarrow 0}\epsilon = 0$. Third,
		Because of Theorem 2, instead of directly maximizing $\mathrm{ELBO}_{\mathrm{EB}}$, we are maximizing the loose low bound $\mathrm{LELBO}_{\mathrm{EB}}$ and then the optimal estimates are
		\begin{align}
		& \hat{\varTheta} = \arg\max\limits_{\varTheta}\lim\limits_{\epsilon \rightarrow 0} \mathrm{LELBO}_{\mathrm{EB}} \nonumber \\
		& = \arg\max\limits_{\varTheta}\lim\limits_{\epsilon \rightarrow 0}
		E_{q(F, U, X, Z)}\log p(Y|F) - \mathrm{KL}(q(X)||p(X)) - \nonumber \\
		& \quad \mathrm{KL}(q(U)||p(U)) - A + B + C \nonumber \\
		& = 
		\arg\max\limits_{\varTheta}\lim\limits_{\epsilon\rightarrow 0} E_{q(F, U, X, Z)}\log p(Y|F) - \mathrm{KL}(q(X)||p(X)) - \nonumber \\
		& \quad \mathrm{KL}(q(U)||p(U)) - A  \nonumber \\  
		& = \arg\max\limits_{\varTheta} E_{q(F, U, X)}\log p(Y|F, Z=\bm \nu) - \mathrm{KL}(q(X)|| p(X)) - \nonumber \\ 
		& \quad\mathrm{KL}(q(U) ||p(U )) - A \nonumber 
		\end{align}
		Due to Lemma 2, this optimization is equivalent to maximizing $\mathrm{ELBO} - M\mathrm{KL}(\hat{q}_z||\hat{q}_x)$ which is exactly the objective function in the regularization framework. Finally, due to Lemma 1, $\bm z_m$ in empirical Bayesian model converges to $\bm \nu$ and then it suggests that optimized $\bm z_m$ in empirical Bayesian model is the same optimized inducing inputs $\bm z_m$ that maximize $\mathrm{MELBO}$.
	}
\end{proof}


\nocite{langley00}

\bibliography{main}
\bibliographystyle{icml2021}